\documentclass{llncs}
\usepackage{multirow}
\usepackage{makeidx}  

\usepackage{makeidx}
\usepackage{graphicx}
\usepackage{amssymb}
\usepackage{amsmath}
\usepackage{amscd}
\usepackage{tikz}
\usetikzlibrary{matrix,arrows}
\usepackage{multicol}

\newcommand{\commentout}[1]{}

\newcommand{\R}{\mathbb{R}}                    

\newcommand{\cbrace}[1]{\mathop{\left\{ #1 \right\}}}
\newcommand{\bracket}[1]{\mathop{\left[ #1 \right]}}

\DeclareMathOperator{\bd}{bd}                  
\DeclareMathOperator{\id}{id}                  
\DeclareMathOperator{\interior}{int}           

\renewcommand{\S}[1]{{\mathcal{#1}}}           
\def\vec#1{\mathchoice{\mbox{\boldmath$\displaystyle#1$}}
{\mbox{\boldmath$\textstyle#1$}}
{\mbox{\boldmath$\scriptstyle#1$}}
{\mbox{\boldmath$\scriptscriptstyle#1$}}}

\newcommand{\widebar}[1]{\overline{\!#1}}   

{%
\end{array}\right.}

\newcounter{algorithm_counter}
{
\\[-1.5ex]
\rule{\textwidth}{\arrayrulewidth}
\end{footnotesize}
\end{minipage}
\setlength{\parindent}{\parindent}
}

{
\\[-1.5ex]
\rule{\textwidth}{\arrayrulewidth}
\end{footnotesize}
\end{minipage}
\setlength{\parindent}{\parindent}
}

\begin{document}
\frontmatter          
\pagestyle{headings}  
\addtocmark{} 
\title{Flip-Flop Sublinear Models for Graphs\\ 
Supplementary Material\\
Proof of Theorem 1}
\titlerunning{Supplementary Material -- Proof of Theorem 1}  
%
\author{Brijnesh Jain}
\authorrunning{Brijnesh Jain} 
%
\tocauthor{Brijnesh Jain}
\institute{Technische Universit\"at Berlin, Germany\\
\email{brijnesh.jain@gmail.com}}

\maketitle              

\begin{abstract}
We prove that there is no class-dual for almost all sublinear models on graphs. 
\end{abstract}

\section{Introduction}
We prove the following result:
\begin{theorem}
There is no class-dual of a sublinear function with probability one.
\end{theorem}
The proof is based on material presented in 

\bigskip

\noindent
[1] B. Jain. \emph{Flip-Flop Sublinear Models for Graphs}, S+SSPR 2014. 

\bigskip

The content of [1] is not included in this contribution. Section 1 introduces the formalism necessary to derive the proof. Section 2 proves auxiliary results. Finally, the proof of Theorem 1 is presented in Section 3.

\section{Preliminaries}
Let $\S{G}$ be a permutation group and $\S{X_G}$ be the quotient of the group action $\S{G}$ on the matrix space $\S{X} = \R^{n \times n}$, where $n$ is some number not less than the number of vertices of the largest graphs. By $\pi$ we denote the natural projection from $\S{X}$ to $\S{X_G}$. For details we refer to \cite{Jain2009,Jain2012}.  The stabilizer of $\vec{w} \in \S{X}$ is a subgroup defined by
\[
\S{G}_{\vec{w}} = \cbrace{\gamma \in \S{G} \,:\, \gamma \vec{w} = \vec{w}}.  
\]
If $\S{G}_{\vec{w}}$ is the trivial group, then each element of the orbit $[\vec{w}]$ has a trivial stabilizer. A graph $X$ is said to be regular, if there is a representation $\vec{x} \in X$ with trivial stabilizer $\S{G}_{\vec{x}}$.

Suppose that $f(X) = W \cdot X + b$ is a sublinear function. A cross section with basepoint $\vec{w} \in \S{X}$ is an injective map $\phi: \S{X_G} \to \S{X}$ satisfying
\begin{enumerate}
\item $W \cdot X = \phi(W)^T\phi(X)$
\item $\pi \circ \phi (X) = X$
\end{enumerate}
for all $X \in \S{X_G}$. We may regard the map $\phi$ as an isometric embedding of the graph space into some Euclidean space with respect to $\vec{w} = \phi(W)$. Note that $\phi$ is not uniquely determined, even for fixed $\vec{w} = \phi(W)$.  The closure of the image $\phi(\S{X_G})$ is the Dirichlet (fundamental) domain with basepoint (center) $\vec{w}$ defined by
\[
\S{D}_{\vec{w}} = \cbrace{\vec{x} \in \S{X} \,:\, \vec{w}^T\vec{x} \geq \widetilde{\vec{w}}^T\vec{x}, \, \widetilde{\vec{w}} \in W}.
\]
A Dirichlet domain is a convex polyhedral cone of dimension $\dim(\S{X})$  with the following properties: (1) $\S{D}_{\vec{w}} $ is well-defined, (2) $\vec{x} \in \S{D}_{\vec{w}}$ iff $(\vec{x}, \vec{w})$ is an optimal alignment, (3) $\pi(\S{D}_{\vec{w}}) = \S{X}_G$, and (4) $\pi$ is injective on the interior of $\S{D}_{\vec{w}}$ \cite{Jain2012}.

\section{Auxiliary Results}

\begin{lemma}\label{lemma:gamma-phi}
Let $\phi : \S{X_G}\rightarrow \S{X}$ be a fundamental cross section with basepoint $\vec{w}$. Then $\gamma \circ \phi$ is a fundamental cross section with basepoint $\gamma \vec{w}$ for all $\gamma \in \S{G}$.
\end{lemma}

\proof
From the axioms of a group action follows that $\gamma$ is bijective. A cross section $\phi$ is injective by definition. Then the composition $\phi' = \gamma \circ \phi$ is also injective. We show that $W \cdot X = \phi'(W)^T\phi'(X)$. We have
\begin{align*}
\phi'(W)^T\phi'(X) = \gamma \vec{w}^T\gamma \vec{x},
\end{align*}
where $\vec{w} = \phi(W)$ and $\vec{x} = \phi(X)$. Since $\S{G}$ is a permutation group, the mapping $\gamma$ is orthogonal. As an orthogonal mapping, $\gamma$ preserves the inner product, that is $\gamma \vec{w}^T\gamma \vec{x} = \vec{w}^T\vec{x}$. Since $\phi$ is a fundamental cross section with basepoint $\vec{w}$, we have
\[
W \cdot X = \phi(W)^T\phi(X) = \vec{w}^T\vec{x} = \gamma \vec{w}^T\gamma \vec{x} = \phi'(W)^T\phi'(X).
\]
Note that this part also shows that the $\phi'$ is a cross section with basepoint $\gamma \vec{w}$. 
Finally, we show that $\pi \circ \phi' (X) = X$ for all $X \in \S{X_G}$. Since $\phi$ is a fundamental cross section, the vector $\vec{x} = \phi(X)$ projects to $X$ via $\pi$. The graph $X$ can be regarded as the orbit $\bracket{\vec{x}}$ of $\vec{x}$. As an element of $\bracket{\vec{x}}$ the vector $\gamma \vec{x}$ also projects to $X$ via $\pi$. This shows $\pi \circ \phi' (X) = X$. 
\qed

\bigskip

\begin{lemma}
Let $\phi : \S{X_G}\rightarrow \S{X}$ be a fundamental cross section  with basepoint $\vec{w}$ and Dirichlet domain $\S{D}_{\vec{w}}$. Then $\vec{w}$ is an interior point of  $\S{D}_{\vec{w}}$ if and only if $W = \pi(\vec{w})$ is regular.
\end{lemma}

\proof
Let $\vec{w}$ be an interior point of $\S{D}_{\vec{w}}$. We assume that $W$ is not regular. Then there is a $\gamma \in \S{G} \setminus \cbrace{\id}$ such that $\vec{w} = \gamma \vec{w}$. According to Lemma \ref{lemma:gamma-phi}, the composition $\phi' = \gamma \circ \phi$ is a fundamental cross section  with basepoint $\gamma\vec{w}$ and Dirichlet domain $\gamma \S{D}_{\vec{w}} = \S{D}_{\gamma\vec{w}}$. From $\vec{w} = \gamma \vec{w}$ follows that 
\[
\S{D} = \interior\S{D}_{\vec{w}} \cap \interior \gamma\S{D}_{\vec{w}} \neq \emptyset,
\]
where $\interior \S{S}$ denotes the interior of a set $\S{S} \subseteq \S{X}$. As an intersection of open convex sets, the set $\S{D}$ is also open and convex. Then there are $n + 1$ points $\vec{x}_0, \ldots, \vec{x}_n \in \S{D}$ in general position, where $n = \dim(\S{X})$.  By definition of a Dirichlet domain, we have
\[
\vec{w}^T\vec{x}_i = \vec{w}^T \gamma \vec{x}_i
\]
for all $i \in \cbrace{0, \ldots, n}$. This implies $\vec{x}_i = \gamma\vec{x}_i$, because cross sections are injective.  Observe that $\S{G}$ acts isometrically on $\S{X}$. Since two isometries are the same if they coincide on $n+1$ points in general position, we obtain $\gamma = \id$. This contradicts our assumption that there is a $\gamma \neq \id$ such that $\vec{w} = \gamma \vec{w}$. Hence, $W$ is regular.

Now we assume that $W$ is regular and show that there is a representation $\vec{w}$ of $W$ such that $\vec{w}\in\interior \S{D}_{\vec{w}}$. The boundary of $\S{D}_{\vec{w}}$ is of the form
\[
\bd \S{D}_{\vec{w}} = \bigcup_{\gamma\in \S{G}\setminus \cbrace{\id}} \S{D}_{\vec{w}} \cap \gamma \S{D}_{\vec{w}}.
\]
Since $W$ is regular, $\gamma \vec{w} \neq \vec{w}$ for all $\gamma \in \S{G}\setminus \cbrace{\id}$

Suppose that $\tilde{\vec{w}} \neq \vec{w}$ is another representation of $W$ with Dirichlet domain $\S{D}_{\tilde{\vec{w}}}$. The bisection of $\S{D}_{\vec{w}}$ and $\S{D}_{\tilde{\vec{w}}}$ is defined by the set 
\[
\S{H}(\vec{w}, \tilde{\vec{w}}) = \cbrace{\vec{x} \in \S{D} \,:\, \vec{w}^T\vec{x} = \tilde{\vec{w}}^T\vec{x}},
\]
where $\S{D} = \S{D}_{\vec{w}} \cap \S{D}_{\tilde{\vec{w}}}$. Since $\vec{w}$ and $\tilde{\vec{w}}$ are unequal, $\S{H}(\vec{w}, \vec{w}')$ is a subset of a hyperplane $\S{H}$ defined by the equation $h(\vec{x}) = (\vec{w} - \tilde{\vec{w}})^T\vec{x}$. The hyperplane $\S{H}$ is perpendicular to the vector $\vec{v} = \vec{w}- \tilde{\vec{w}}$  and passes through the midpoint of the connecting line between $\vec{w}$ and $\tilde{\vec{w}}$. This shows that $\vec{w}$ is not a point on $\S{H}(\vec{w}, \tilde{\vec{w}})$. Since $\tilde{w}$ was chosen arbitrarily,  $\vec{w}$ is in the interior of $\S{D}_{\vec{w}}$.
\qed

\bigskip

\begin{lemma}\label{lemma:-w}
Let $W \in \S{X_G}$ be regular. Suppose that $\phi$  is a fundamental cross section with basepoint $\vec{w}$ and Dirichlet domain $\S{D}_{\vec{w}}$. Then $-\vec{w} \notin \S{D}_{\vec{w}}$. 
\end{lemma}

\proof
Observe that $\vec{w} \neq \vec{0}$ and $\vec{w} \neq -\vec{w}$ for all representations $\vec{w}$ of a regular graph $W$. In addition, with $W$ the graph $W' = \pi(-\vec{w})$ is also regular. As shown in \cite{Jain2009}, we have
\begin{align}\label{eq:lemma:-w:01}
W \cdot W = \max_{\gamma \in \S{G}} \;\gamma \vec{w}^T \vec{w} = \vec{w}^T\vec{w}.
\end{align}
Let $W'$ be the graph represented by $-\vec{w}$. From eq.\ \eqref{eq:lemma:-w:01} follows
\begin{align}\label{eq:lemma:-w:02}
-\vec{w}^T\vec{w} = -\min_\gamma \;\gamma \vec{w}^T \vec{w} < \max_\gamma -\vec{w}^T\vec{w} = W' \cdot W
\end{align}
Strict inequality in eq.\ \eqref{eq:lemma:-w:02} follows from regularity of $W$ and $W'$ together with $\vec{w} \neq -\vec{w}$. Thus, $-\vec{w}$ is not an element of the Dirichlet domain $\S{D}_{\vec{w}}$ with basepoint $\vec{w}$. 
\qed


\section*{Proof of Theorem 1}
Suppose that each graph $X \in \S{X_G}$ has a class label $y \in \S{Y} = \cbrace{\pm 1}$. By $\mathbb{C}[f]$ we denote the expected misclassification error of the sublinear function $f$. Consider a sublinear function of the form $f(X) = W \cdot X + b$, where $W$ is regular and $b\neq 0$. The equation $f(X) = 0$ defines a decision surface $\S{H}_f$ that separates the graph space $\S{X_G}$ into two disjoint regions $\S{R}_+(f)$ and $\S{R}_-(f)$. By construction we have $\mathbb{C}[f] = 0$.

Let $\phi$ be a fundamental cross section with basepoint $\vec{w} = \phi(W)$ and Dirichlet domain $\S{D}_{\vec{w}}$. By $\S{D}_+(f) = \phi(\S{R}_+(f))$ and $\S{D}_-(f) = \phi(\S{R}_-(f))$ we denote the images of both class regions  $\S{R}_+(f)$ and $\S{R}_-(f)$ in $\S{D}_{\vec{w}}$. The hyperplane separating $\S{D}_+(f)$ and $\S{D}_-(f)$ is defined by the equation $h(\vec{x}) = \vec{w}^T\vec{x} + b = 0$. By construction, the expected misclassification error of $h(\vec{x})$ is $\mathbb{C}_{\S{X}}[h] = 0$. In addition, $h(\vec{x})$ is the unique global minimum of $\mathbb{C}_{\S{X}}[\cdot]$ over all linear functions. 

Now we relabel both class regions $\S{R}_+(f)$ and $\S{R}_-(f)$ such that all graphs from the positive class region  $\S{R}_+(f)$ have negative labels and all graphs from the negative class region $\S{R}_-(f)$ have positive labels. We denote the relabeled regions in $\S{X_G}$ by $\S{\widebar{R}}_+(f)$ and $\S{\widebar{R}}_-(f)$, resp., and similarly the relabeled regions in $\S{D}_{\vec{w}}$ by $\S{\widebar{D}}_+(f)$ and $\S{\widebar{D}}_-(f)$. For the relabeled variant, the linear classifier in $\S{D}_{\vec{w}}$ determined by $h'(\vec{x}) = -\vec{w}^T\vec{x} + b$ has also minimal misclassification error  $\mathbb{C}_{\S{X}}[h'] = 0$ and is the unique minimizer of  $\mathbb{C}_{\S{X}}[\cdot]$. By Lemma \ref{lemma:-w} the opposite direction $-\vec{w}$ of basepoint $\vec{w}$ is not an element of the Dirichlet domain $\S{D}_{\vec{w}}$.

Suppose that $f'(X) = W' \cdot X + b'$ is a sublinear function such that 
\[
\mathbb{C}[f'] = \min_{g} \mathbb{C}[g]
\]
under the relabeled setting with class regions $\S{\widebar{R}}_+(f)$ and $\S{\widebar{R}}_-(f)$. Let $\vec{w}' = \phi(W')$ be a representation of $W'$ in $\S{D}_{\vec{w}}$ and let $\phi'$ be a fundamental cross section with basepoint  $\vec{w}' = \phi'(W')$ and Dirichlet domain $\S{D}_{\vec{w}'}$. We consider the intersection $\S{D} = \S{D}_{w} \cap \S{D}_{\vec{w}'}$. According to \cite{Gerstenhaber1951}, Theorem 12, the intersection of two convex polyhedral cones is again a convex polyhedral cone. We show that $\S{D}$ has co-dimension $0$. For this, we first show that the relative interior of the intersection $\S{D}$ contains an inner point $\vec{z}$ from $\S{D}_{\vec{w}}$. Since $W$ is regular, $\vec{w}$ is an inner point of $\S{D}_{\vec{w}}$ lying in $\S{D}$. Two cases can occur: (i) $\vec{w}$ is in the relative interior of $\S{D}$; and (ii) $\vec{w}$ lies on the boundary of $\S{D}$. If (i) holds, we set $\vec{z} = \vec{w}$. Suppose that $(ii)$ holds. Since $\vec{w}$ and $\vec{w}'$ are distinct elements of the intersection $\S{D}$, any point $\vec{z} = \lambda \vec{w} + (1-\lambda)\vec{w}'$ with $0 < \lambda < 1$ lies in the relative interior of $\S{D}$ by convexity. 

Next we show that there is an $\varepsilon > 0$ such that the open ball $\S{B}(\vec{z}, \varepsilon)$ with center $\vec{z}$ and radius $\varepsilon$ is contained in $\S{D}$. We choose $\varepsilon < \min(\lambda, 1- \lambda)$. Then $\S{B}(\vec{z}, \varepsilon) \subset \S{D}_{\vec{w}}$, because $\vec{z}$ is an interior point of $\S{D}_{\vec{w}}$. Suppose that $\S{B}(\vec{z}, \varepsilon)$ is not contained in $\S{D}$. Since $\vec{z}$ lies in the relative interior of $\S{D}$, the co-dimension of $\S{D}$ is positive. In addition, $\vec{z}$ lies on the boundary of $\S{D}_{\vec{w}'}$. This implies that $\vec{z}$ is also a boundary point of $\S{D}_{\vec{w}}$, which is a contradiction of our construction. Hence, the co-dimension of $\S{D}$ is zero.

Since $b \neq 0$, the hyperplane defined by equation $h(\vec{x}) = \vec{w}^T\vec{x} + b = 0$ does not pass through the origin $\vec{0}$. Boundaries of any Dirichlet domain are supported by hyperplanes passing through $\vec{0}$. Thus the intersection $\S{D}$ contains an open set $\S{U}$ separated by the hyperplane segment $\S{H}_f$. Then there are $n+1$ points $\vec{x}_{0}, \ldots, \vec{x}_n \in \S{D}$ in general position that fix the unique hyperplane.  Since $-\vec{w} \notin \S{D}_{\vec{w}}$, we have $\mathbb{C}[f'] > 0$ implying that $f'$ is not a dual of $f$. 

The probabilistic statement follows from the fact that interior points are regular and therefore have Lebesgue measure one. With respect to the bias $b$, the set $\R\setminus\cbrace{0}$ also has Lebesgue measure one. Combining all parts we arrive at the assertion. 
\qed

%
%
%


\begin{thebibliography}{}
%
\bibitem{Jain2009}
B. Jain and K. Obermayer.
\newblock Structure Spaces.
\newblock \emph{The Journal of Machine Learning Research}, 10:2667--2714, 2009.

\bibitem{Jain2012}
B. Jain and K. Obermayer.
\newblock Learning in Riemannian Orbifolds.
\newblock \emph{arXiv preprint arXiv:1204.4294}, 2012.

\bibitem{Gerstenhaber1951}
M. Gerstenhaber. 
\newblock Theory of convex polyhedral cones. 
\newblock \emph{Activity analysis of production and allocation }, 298--316, 1951.
\end{thebibliography}
\end{document}